\documentclass[letterpaper,10pt,journal,twoside]{IEEEtran}

\usepackage{amsmath,amssymb,amsfonts,bm}
\usepackage{graphicx}
\usepackage{booktabs}
\usepackage{multirow}
\usepackage{array}
\usepackage{xcolor}
\usepackage{cite}
\usepackage{url}
\usepackage[hidelinks]{hyperref}
\usepackage{capt-of}
\usepackage{kotex}
\usepackage{tabularx}

\newcommand{\Ts}{T_{s}}

\hypersetup{
  pdftitle={Simple Torque-Observation Alignment for Zero-Shot Sim-to-Real Grasping with a Direct-Drive Gripper},
  pdfauthor={Doyoung Kim, Edgar Lee, Hyeonsun Park, Chunghyeon Lee, Chihyun Han, Uisu Hwang, Seokhwan Jeong}
}

\begin{document}
\pagestyle{plain}
\title{Simple Torque-Observation Alignment for Zero-Shot
Sim-to-Real Grasping with a Direct-Drive Gripper}

\author{Doyoung~Kim, Edgar~Lee, Hyeonsun~Park, Chunghyeon~Lee,
Chihyun~Han, Uisu~Hwang, and~Seokhwan~Jeong$^{*}$%
\thanks{All authors are with the Department of Mechanical Engineering,
Sogang University, Seoul, South Korea.
$^{*}$Corresponding author: Seokhwan Jeong
(e-mail: \href{mailto:seokhwan@sogang.ac.kr}{seokhwan@sogang.ac.kr}).}}


\IEEEaftertitletext{%
  \vspace{-6mm} 

  \begin{minipage}{\textwidth}
    \centering
    \includegraphics[
      width=1.0\textwidth,
      keepaspectratio
    ]{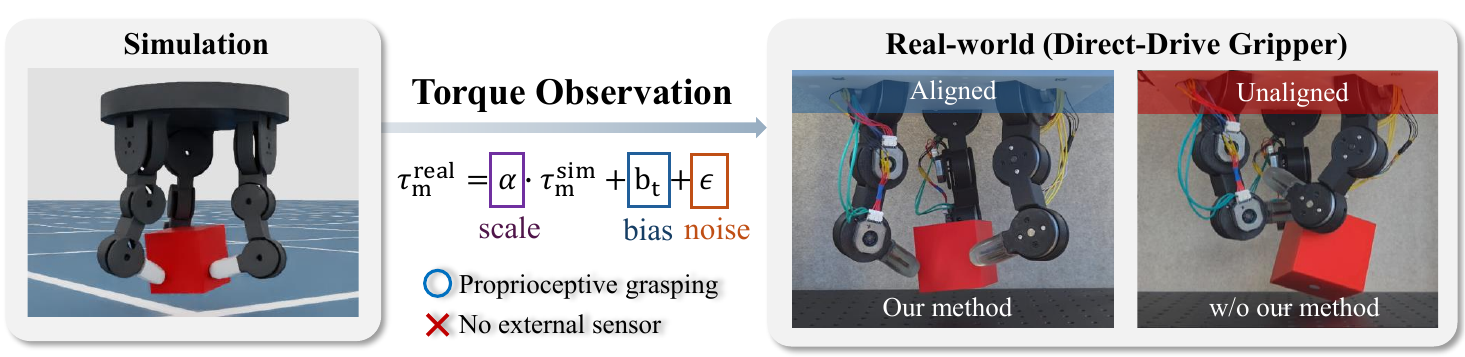}

    \vspace{-2.5mm} 
    \captionof{figure}{Overview of the proposed simple torque-observation alignment method for zero-shot sim-to-real grasping with a direct-drive gripper.}
    \label{fig:framework}
  \end{minipage}

  \vspace{1.5mm} 
}

\maketitle
\thispagestyle{plain}

\begin{abstract}
Torque observations in reinforcement learning remain challenging because simulated and measured torque differ in scale, offset, and noise.  In this paper, we propose a simple torque observation alignment method for robots with direct-drive (DD) actuators, in which motor current maps linearly
to joint torque through a \textcolor{black}{motor-type-specific} torque constant $K_\tau$. First, dynamometer calibration identifies $K_\tau^\ast$ and corrects the scale mismatch between simulated and real torque. Second, the method uses $\Delta\tau_t=\tau_t-\tau_{t-1}$ as the observation in both domains to eliminate the \textcolor{black}{constant} offset instead of using the direct torque $\tau_t$, which carries a domain-dependent bias. Third, Gaussian noise obtained from the dynamometer measurement data is injected
during the learning process. To validate the proposed method, we train a teacher--student grasping
policy entirely in simulation and deploy the distilled student on a multifingered DD gripper. The deployed policy performs proprioceptive grasping using
only joint positions and torque differences. We conduct an ablation study comparing the proposed method with
alternative alignment variants on nine in-distribution (ID) objects. The proposed method achieves
100\% grasp success. These results demonstrate that the proposed alignment method improves
the robustness of zero-shot policy transfer on the DD gripper against
real-world torque-observation mismatches.
\end{abstract}


\section{Introduction}
\IEEEPARstart{S}{imulation-based} reinforcement learning has become a widely used
approach to contact-rich robotic manipulation as it can generate
large amounts of interaction experience without costly and
time-consuming hardware trials~\cite{openai2020,dextreme2023,chen2021reorientation,zhang2025robustdexgrasp,bauza2024demostart,lum2026play2perfect}. \textcolor{black}{Grasping requires information about both the
gripper and the object, yet the object's state and properties
are often not fully known. Feedback from physical interaction
can help address this uncertainty, and joint torque can
complement joint-position feedback by providing} proprioceptive information about contact onset and changes
in grasp load.

In direct-drive (DD) actuation, torque is transmitted directly
to the robot joints without a gearbox~\cite{bhatia2019direct}. The gearless configuration reduces friction
and backlash and promotes backdrivability and high mechanical transparency. This allows
motor-side proprioceptive signals to reflect changes in interaction
loads \textcolor{black}{more directly than in geared actuators,
where high reflected inertia,  transmission friction and backlash can obscure
these changes~\cite{bhatia2019direct}}. Together with an approximately linear
current-to-torque relationship, these characteristics make motor
current a practical source of torque observations for learning-based
contact-rich manipulation.

\par However, even in robotic systems with DD actuators, simulated and real
joint-torque observations can differ in scale, additive bias,
and noise. In simulation, actuator torque is computed from the control
command and the simulated dynamics~\cite{todorov2012mujoco,mittal2025isaaclab}, whereas the real-world torque
observation considered here is estimated from motor current.
Scale mismatch can arise from errors in the torque constant or
current calibration used for current-to-torque conversion~\cite{corke1996insitu,lin2009torqueconstant}.
Bias mismatch occurs when simulated and real-world torque observations
contain constant or slowly varying offsets, resulting in
different observation baselines~\cite{hu2020offset}.
Noise mismatch arises from effects in real-world torque measurements,
such as electrical fluctuations and quantization, that are not modeled
identically in simulation~\cite{hu2020offset}.
\textcolor{black}{\textcolor{black}{These discrepancies can cause a
simulation-trained policy to interpret contact and load
changes differently in the real world. Even with the high
mechanical transparency of DD actuators, torque-observation
mismatch therefore remains a challenge that needs to be
addressed for reliable policy transfer.}}

\begin{figure*}[t]
    \centering
    \includegraphics[width=\textwidth]{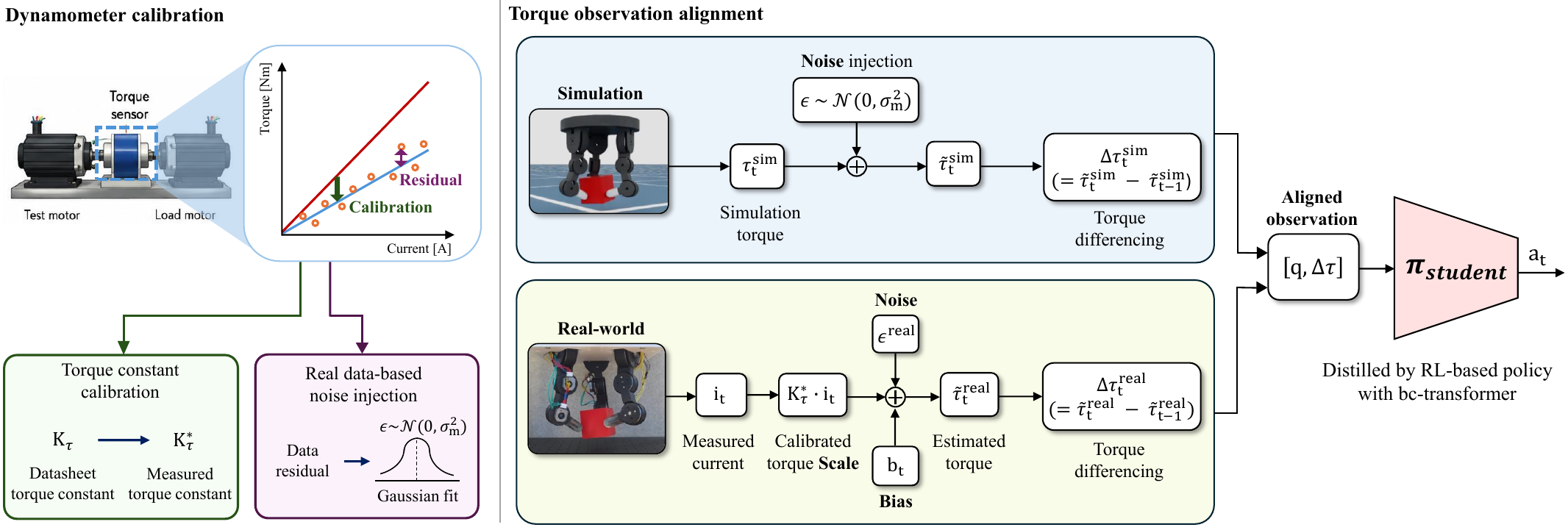}
    \vspace{-7mm}
    \caption{Simple torque observation alignment method pipeline.}
    \label{fig:pipeline}
    \vspace{-0.4cm}
\end{figure*}

Prior work addresses torque-related sim-to-real discrepancies through
implicit and explicit methods. Implicit methods improve policy
robustness to domain differences through dynamics randomization and
torque perturbations~\cite{erfi2022,cha2025}. Such training increases
tolerance to modeling errors without directly matching simulated and
real torque signals. However, these methods do not explicitly
establish a calibrated correspondence between simulated and real
torque observations, and their effectiveness depends on the chosen
randomization ranges. Explicit methods directly use real-world measurements to reduce the mismatch between the two domains. Actuator-modeling methods reproduce
real actuator behavior in simulation using analytical models or learned
networks~\cite{tan2018,masuda2022,hwangbo2019,rudin2022}. This
real-to-simulation mapping improves simulation fidelity. However,
actuator modeling requires actuator-specific identification or model
training. Observation-alignment methods instead convert real torque
measurements to match the policy input used in simulation
~\cite{singh2023bipedal,bytedance2026}. \textcolor{black}{Closely related work~\cite{bytedance2026} aligns normalized motor
current with simulated torque observations through fingertip contact-force measurements and uses tactile feedback during grasping.
Our method combines motor-type-specific dynamometer calibration
with temporal torque differencing, enabling deployment using only
joint positions and motor-current-derived torque changes, without tactile sensors.} These considerations motivate a simple torque-observation method for
robots with DD actuators that combines explicit cross-domain alignment with implicit robustness training.

We propose a simple torque-observation alignment method between the
simulation and real-world domains for robots with DD actuators (Fig.~\ref{fig:framework}). First, we apply a dynamometer calibration method to identify the torque constant $K_\tau^\ast$. $K_\tau^\ast$ converts
measured motor current into a torque estimate, reducing the
cross-domain scale mismatch. \textcolor{black}{For the two motor types used in our DD gripper, datasheet-based
torque estimates exceed dynamometer-calibrated estimates by
approximately 38.0\% and 7.8\% (Table~\ref{tab:kt}).
These unequal errors distort the relative torque pattern across
joints and motivate motor-type-specific observation alignment.} Second, we use the torque difference $\Delta\tau_t=\tau_t-\tau_{t-1}$ \textcolor{black}{as policy input} in both domains to cancel constant offsets and attenuate slowly varying baselines. Torque-change observations provide proprioceptive cues about changing interaction loads with reduced sensitivity to observation baselines. Third, we add Gaussian noise to simulated torque observations
during training, using dynamometer calibration data to set
the noise level. The proposed method thus provides a compact observation-processing
pipeline for zero-shot sim-to-real policy transfer.

We validate the proposed method through proprioceptive grasping
with a multifingered DD gripper. Trained entirely in simulation, the deployed policy
grasps, lifts, and holds objects using only actuator proprioception (joint position $q_t$ and joint-torque change $\Delta\tau_t$),
without vision or dedicated contact sensors.
Real-world ablations show the benefit of the proposed alignment,
while tests on unseen objects demonstrate transfer beyond the
training object set.

The main contributions of this work are as follows:

\begin{itemize}
    \item We present a framework for learning proprioceptive
    grasping policies for a direct-drive multifingered gripper.
    By integrating torque-observation alignment with
    simulation-based teacher--student learning, the framework
    enables zero-shot sim-to-real transfer using only joint
    positions and torque differences, without vision or
    dedicated tactile or force sensors during deployment.

    \item We approximate torque-observation mismatch using a
    physically motivated affine model with a motor-type-specific
    scale factor, a constant or slowly varying offset, and
    measurement noise. The method combines dynamometer-based
    torque-constant calibration, temporal differencing, and
    measurement-informed Gaussian noise injection without
    learning a nonlinear mapping.

    \item We validate the framework through real-world
    ablations and zero-shot grasp--lift--hold experiments.
    Under the evaluated conditions, the proposed method
    achieves 100.0\% success on nine in-distribution objects
    and 99.0\% overall across 21 objects, including twelve
    previously unseen out-of-distribution objects.
\end{itemize}
\par A \textit{supplementary video} accompanying this paper demonstrates real-world grasping, lifting, and holding with the proposed method.

\section{Method}\label{sec:method}
An overview of the proposed method is shown in Fig.~\ref{fig:pipeline}. It has two parts. First, we align the joint-torque observation across domains (Section~\ref{sec:align}). Second, we train a teacher policy for the grasping task in simulation
using privileged information and the aligned torque observation, and
distill it into a student policy for real-world deployment
(Section~\ref{sec:learn}).
\subsection{Direct-Drive Gripper Platform}\label{sec:hardware}

\begin{figure}[t]
    \centering
    \includegraphics[width=\columnwidth]{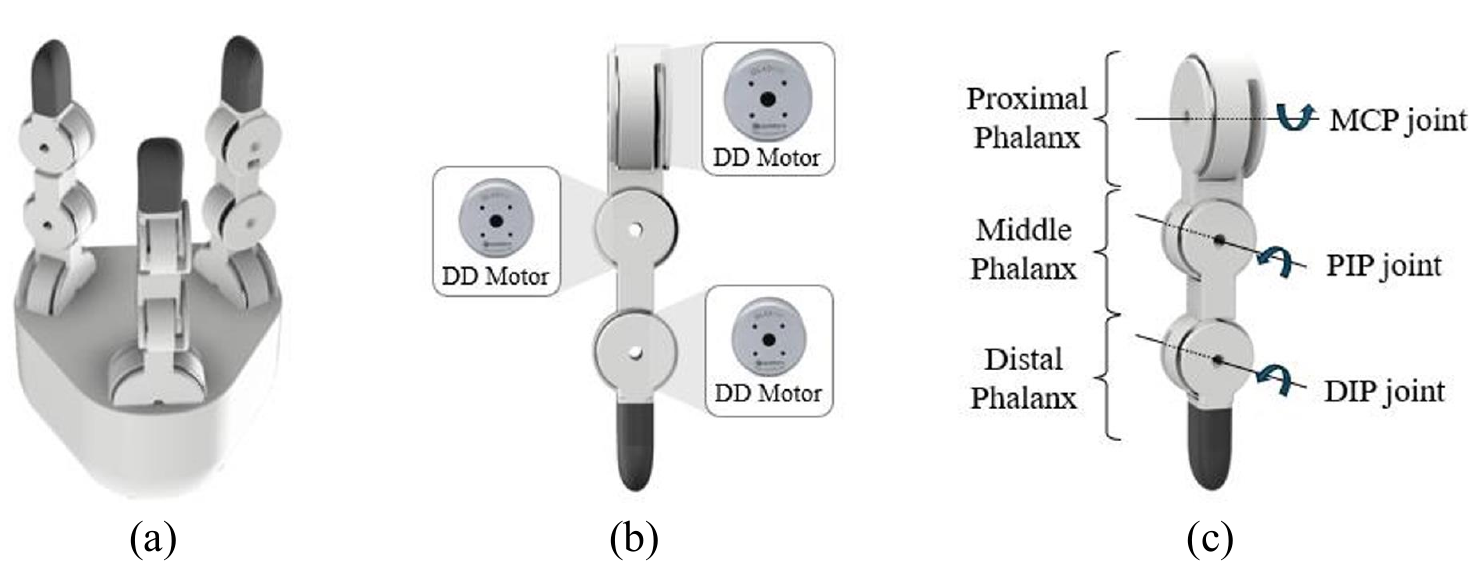}
    \vspace{-7mm}
    \caption{Direct-drive gripper platform. (a) DD actuator-based multifingered gripper.
    (b) Direct-Drive (DD) actuators integrated into a single finger. (c) Joint and link
    configuration.}
    \label{fig:gripper_platform}
    \vspace{-0.35cm}
\end{figure}

\textcolor{black}{The hardware platform is a three-fingered, 9-DoF gripper
(see Fig.~\ref{fig:gripper_platform}). Each finger has three independently
actuated joints, with a DD motor integrated coaxially into each joint
without a mechanical transmission. The metacarpophalangeal (MCP) joint
is driven by a 69-mm-diameter DD motor (CubeMars, GL60 KV25), whereas the
proximal interphalangeal (PIP) and distal interphalangeal (DIP) joints
are each driven by a 46.5-mm-diameter DD motor (CubeMars, GL40 KV70).
A microcontroller unit (ROBOTIS, OpenCR 1.0) acquires joint positions
and quadrature-axis currents $i_q$ from brushless motor controllers
(ODrive Robotics, ODrive Micro) over CAN.
The measured $i_q$ is converted into a joint-torque estimate using the
corresponding torque constant. The gripper model and grasping task
environment are implemented in NVIDIA Isaac Sim using the Isaac Lab
framework~\cite{mittal2025isaaclab}.}

\subsection{Simple Torque Observation Alignment}\label{sec:align}

\begin{figure}[t]
\centering
\includegraphics[width=\columnwidth]{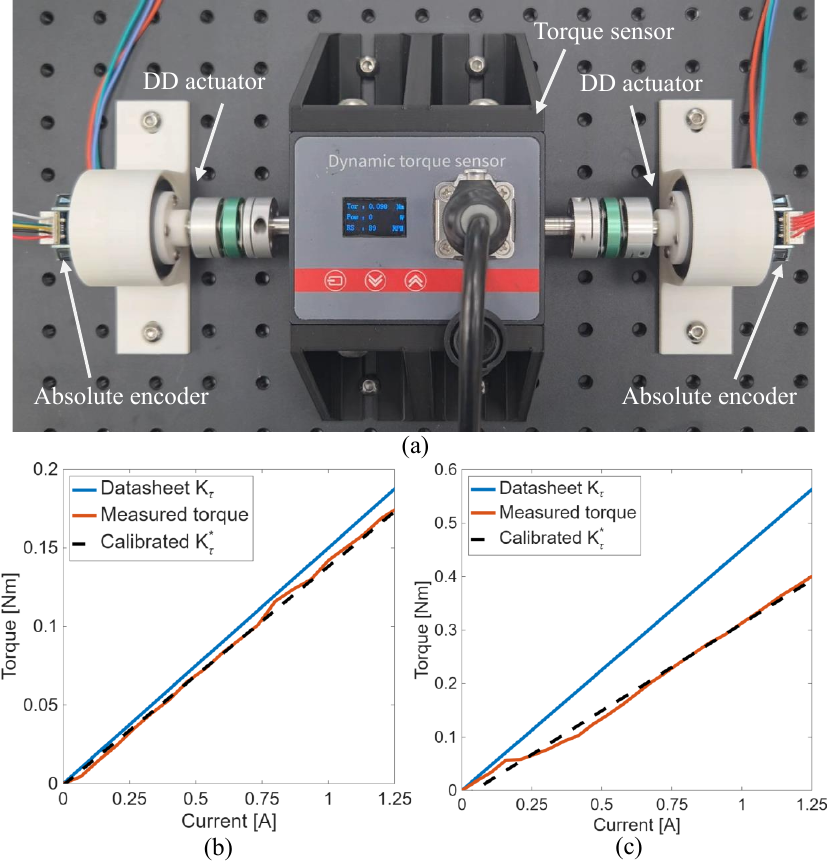}
\vspace{-7mm}
\caption{Dynamometer-based identification of the effective torque
constants. (a) Experimental setup consisting of a test motor, a load
motor, and an inline torque sensor. (b) Measured shaft torque versus
quadrature current for the GL40 motor. (c) The corresponding result for
the GL60 motor. Dashed lines denote the linear regression fits.}
\label{fig:kt}
\vspace{-0.35cm}
\end{figure}

\textbf{Cross-Domain Torque Model.}
The policy receives a joint-torque observation in both domains, although
the observations are generated differently. In simulation, actuator torque is computed by the proportional--derivative
(PD) controller and applied to the simulated joints. In the real world, the torque observation is estimated from the measured
motor current as $\hat{\tau}_{m,t}=K_\tau i_q(t)$, where $K_\tau$ is the torque
constant used for current-to-torque conversion. We represent the cross-domain relation using a multiplicative scale
factor, an additive baseline, and a residual term:
\begin{equation}
\hat{\tau}_{m,t}^{\mathrm{real}}
=
\alpha\tau_{m,t}^{\mathrm{sim}}
+b_t+\epsilon_t,
\qquad
\alpha=\frac{K_\tau}{K_\tau^\ast}.
\label{eq:model}
\end{equation}
We take \eqref{eq:model} as the premise of the method. The scale factor $\alpha$ is the ratio of the torque constant used for
current-to-torque conversion to the calibrated torque constant $K_\tau^\ast$. The term $b_t$ represents a slowly varying baseline mismatch between the
two domains, including contributions from gravity, friction, and current
offset. The term $\epsilon_t$ denotes residual noise from measurements and
unmodeled actuator effects. Overall, \eqref{eq:model}
expresses the real current-derived torque observation as a scaled
simulated actuator torque with a slowly varying baseline and residual
error.

\textbf{Scale: Dynamometer Calibration.}
We reduce the torque-scale mismatch by identifying the effective torque
constant $K_\tau^\ast$ through dynamometer calibration and using it for
real-world current-to-torque conversion.
Differences between nominal and effective torque constants arise from
motor variation, current calibration, and current conventions~\cite{corke1996insitu,lin2009torqueconstant}. This discrepancy causes a
scale mismatch between simulated and real torque observations.

We identify the effective torque constants of the DD actuators
(GL60 KV25 and GL40 KV70) using a motor-loaded rotary
dynamometer~\cite{katz2016dyno}. As shown in Fig.~\ref{fig:kt}, each test motor is coupled coaxially to a load motor through an inline torque sensor \textcolor{black}{(CALT Sensor, DYN-200)}. The controller applies $i_q$ commands while the load motor provides opposing torque. The controller-reported current $i_q$ and measured shaft torque are recorded simultaneously. For each motor type, we linearly regress the measured shaft torque
against the corresponding controller-reported current $i_q$.
The \textcolor{black}{GL60 KV25 motor }
regression uses all recorded operating points up to
$0.500$~Nm, whereas the \textcolor{black}{GL40 KV70 motor } regression is restricted to input torques
not exceeding $0.300$~Nm. Table~\ref{tab:kt} summarizes the identified
torque constants and regression accuracy. Only the fitted slope $K_\tau^\ast$ is used for current-to-torque conversion. 
\begin{table}[!t]
\centering
\caption{Datasheet and dynamometer-identified torque constants with
regression accuracy.}
\label{tab:kt}
\footnotesize
\setlength{\tabcolsep}{3pt}
\renewcommand{\arraystretch}{1.15}
\begin{tabular}{lcccc}
\toprule
\multicolumn{1}{c}{\multirow{2}{*}{Motor}}
& Datasheet $K_\tau$
& Calibrated $K_\tau^\ast$
& \multirow{2}{*}{$R^2$}
& \multirow{2}{*}{RMSE [Nm]} \\
& [Nm/A]
& [Nm/A]
&
& \\
\midrule
GL60 KV25 & 0.4500 & 0.3260 & 0.9940 & 0.0098 \\
GL40 KV70 & 0.1500 & 0.1392 & 0.9992 & 0.0023 \\
\bottomrule
\end{tabular}
\vspace{-0.25cm}
\end{table}
\begin{table}[!t]
\centering
\caption{Motor-type-specific Gaussian noise derived from dynamometer
regression residuals and injected into simulated torque observations.}
\label{tab:residual_noise}
\footnotesize
\setlength{\tabcolsep}{2.5pt}
\renewcommand{\arraystretch}{1.15}
\begin{tabular*}{\columnwidth}{
@{\extracolsep{\fill}}lcc@{}}
\toprule
\multicolumn{1}{c}{Motor}
& Gaussian model $\epsilon_{m,t}$
& Injected std. $\sigma_m$ [Nm] \\
\midrule
GL60 KV25
& $\mathcal{N}(0,9.688465{\times}10^{-5})$
& 0.009843 \\
GL40 KV70
& $\mathcal{N}(0,5.313025{\times}10^{-6})$
& 0.002305 \\
\bottomrule
\end{tabular*}
\vspace{-0.25cm}
\end{table}

\textbf{Bias: Temporal Differencing.}
To suppress the torque-observation bias between the two domains, we
apply temporal differencing to both observations:
\begin{equation}
\Delta\tau_{t}=\tau_{m,t}-\tau_{m,t-1}.
\label{eq:delta}
\end{equation}
After dynamometer calibration, applying \eqref{eq:delta} to
\eqref{eq:model} gives
\begin{equation}
\Delta\hat{\tau}^{\mathrm{real}}_{m,t}
=
\Delta\tau^{\mathrm{sim}}_{m,t}
+
\underbrace{(b_t-b_{t-1})}_{\text{baseline residual}}
+
\underbrace{(\epsilon_t-\epsilon_{t-1})}_{\text{differenced residual}}.
\label{eq:diffed}
\end{equation}
Temporal differencing cancels this baseline when $b_t=b_{t-1}$
and attenuates its slowly varying components. It thus reduces sensitivity to observation
baselines while preserving cues about changes in interaction loads.
\begin{figure}[t]
\centering
\includegraphics[width=\columnwidth]{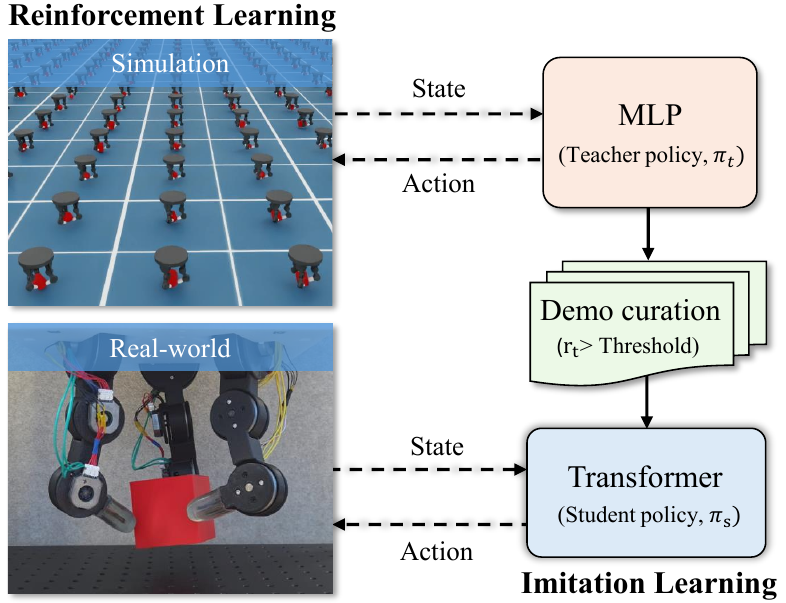}
\vspace{-7mm}
\caption{Teacher-student policy training and deployment pipeline.}
\label{fig:learning_framework}
\vspace{-0.35cm}
\end{figure}

\begin{table}[t]
\centering
\caption{Training hyperparameters.}
\label{tab:training_hyperparameters}
\footnotesize
\setlength{\tabcolsep}{2pt}
\renewcommand{\arraystretch}{1.0}

\begin{minipage}[t]{0.48\columnwidth}
\centering
\begin{tabular*}{\linewidth}{@{\extracolsep{\fill}}lc@{}}
\toprule
\multicolumn{2}{c}{\textbf{Teacher (PPO)}} \\
\midrule
Parameter & Value \\
\midrule
Learning rate            & $10^{-3}$ \\
Steps per environment    & 24 \\
PPO update epochs        & 5 \\
Minibatches              & 6 \\
Activation               & ELU \\
Discount factor $\gamma$ & 0.99 \\
GAE parameter $\lambda$  & 0.95 \\
Entropy coefficient      & 0.003 \\
\bottomrule
\end{tabular*}
\end{minipage}\hfill%
\begin{minipage}[t]{0.48\columnwidth}
\centering
\begin{tabular*}{\linewidth}{@{\extracolsep{\fill}}lc@{}}
\toprule
\multicolumn{2}{c}{\textbf{Student (BC)}} \\
\midrule
Parameter & Value \\
\midrule
Learning rate       & $10^{-4}$ \\
Batch size          & 256 \\
Optimizer           & AdamW \\
Loss function       & MSE \\
Context length      & 30 \\
Transformer layers  & 6 \\
Embedding dimension & 512 \\
Attention heads     & 8 \\
\bottomrule
\end{tabular*}
\end{minipage}
\par\vspace{-3mm}
\end{table}

\begin{table}[t]
\centering
\caption{Dimensions of the training objects.}
\label{tab:object_dimensions}
\footnotesize
\setlength{\tabcolsep}{3pt}
\renewcommand{\arraystretch}{1.05}
\begin{tabular*}{\columnwidth}{
@{\hspace{2mm}\extracolsep{\fill}}llccc@{}
}
\toprule
Shape & Dimensions [cm] & A & B & C \\
\midrule
Cuboid   & Base length $\times$ height
         & $6\times6$ & $8\times8$ & $10\times8$ \\
Cylinder & Radius $\times$ height
         & $4\times4$ & $4.5\times4.5$ & $5\times5$ \\
Sphere   & Radius
         & $4$ & $4.5$ & $5$ \\
\bottomrule
\end{tabular*}
\vspace{-3mm}
\end{table}

\begin{table}[t]
\centering
\caption{Randomization applied during teacher training.}
\label{tab:events}
\footnotesize
\setlength{\tabcolsep}{4pt}
\renewcommand{\arraystretch}{1.1}

\begin{tabularx}{\columnwidth}{
@{}>{\raggedright\arraybackslash}X c@{}
}
\toprule
\textbf{Parameter} & \textbf{Range} \\
\midrule

Robot: Initial MCP Joint Offset (rad)
& $+\mathcal{U}[-0.06,\,0.06]$ \\
Robot: Initial PIP Joint Offset (rad)
& $+\mathcal{U}[-0.35,\,0.35]$ \\
Object: Initial Position ($x$, $y$) (m)
& $+\mathcal{U}[-0.01,\,0.01]$ \\
Object: Initial Yaw (rad)
& $\mathcal{U}[-\pi,\,\pi]$ \\
Object: Mass
& $\times\mathcal{U}[0.5,\,3.0]$ \\
Actuator: P (Stiffness) Gain
& $\times\mathcal{U}[0.85,\,1.05]$ \\
Actuator: D (Damping) Gain
& $\times\mathcal{U}[0.75,\,1.50]$ \\
Static Friction Coefficient
& $\mathcal{U}[0.7,\,1.3]$ \\
Dynamic Friction Coefficient
& $\mathcal{U}[0.7,\,1.3]$ \\
Perturbation: Downward Velocity (m/s)
& $\mathcal{U}[-0.5,\,0]$ \\
Perturbation: Interval (s)
& $\mathcal{U}[8,\,10]$ \\
\bottomrule
\end{tabularx}

\vspace{-0.2cm}
\end{table}

\textbf{Noise: Dynamometer-Residual Gaussian Injection.}
To improve policy robustness to residual noise in real-world torque
observations, we inject Gaussian noise derived from
experimental dynamometer data into the simulated torque observations during teacher training. For each motor type, we use the dynamometer regression RMSE reported in
Table~\ref{tab:kt} as the standard deviation of the injected
Gaussian noise. The resulting noise models are summarized in
Table~\ref{tab:residual_noise}.

Let $\sigma_j$ denote the noise standard deviation assigned to joint
$j$ according to its motor type: $\sigma_j=0.009843$~$\mathrm{N\,m}$
for 
\textcolor{black}{MCP joints} and $\sigma_j=0.002305$~$\mathrm{N\,m}$ for \textcolor{black}{PIP and DIP joints}. At each observation step, Gaussian noise is added
to the simulated torque observation as
\begin{equation}
\widetilde{\tau}^{\mathrm{sim}}_{j,t}
=
\tau^{\mathrm{sim}}_{j,t}
+
\epsilon_{j,t},
\qquad
\epsilon_{j,t}
\sim
\mathcal{N}\!\left(0,\sigma_j^2\right).
\label{eq:noise}
\end{equation}
The policy receives the temporally differenced observation
\begin{equation}
\begin{aligned}
\Delta\widetilde{\tau}^{\mathrm{sim}}_{j,t}
&=
\widetilde{\tau}^{\mathrm{sim}}_{j,t}
-
\widetilde{\tau}^{\mathrm{sim}}_{j,t-1}\\
&=
\Delta\tau^{\mathrm{sim}}_{j,t}
+
\epsilon_{j,t}
-
\epsilon_{j,t-1}.
\end{aligned}
\label{eq:noisy_delta}
\end{equation}

\begin{figure*}[!t]
\centering
\includegraphics[width=\textwidth]{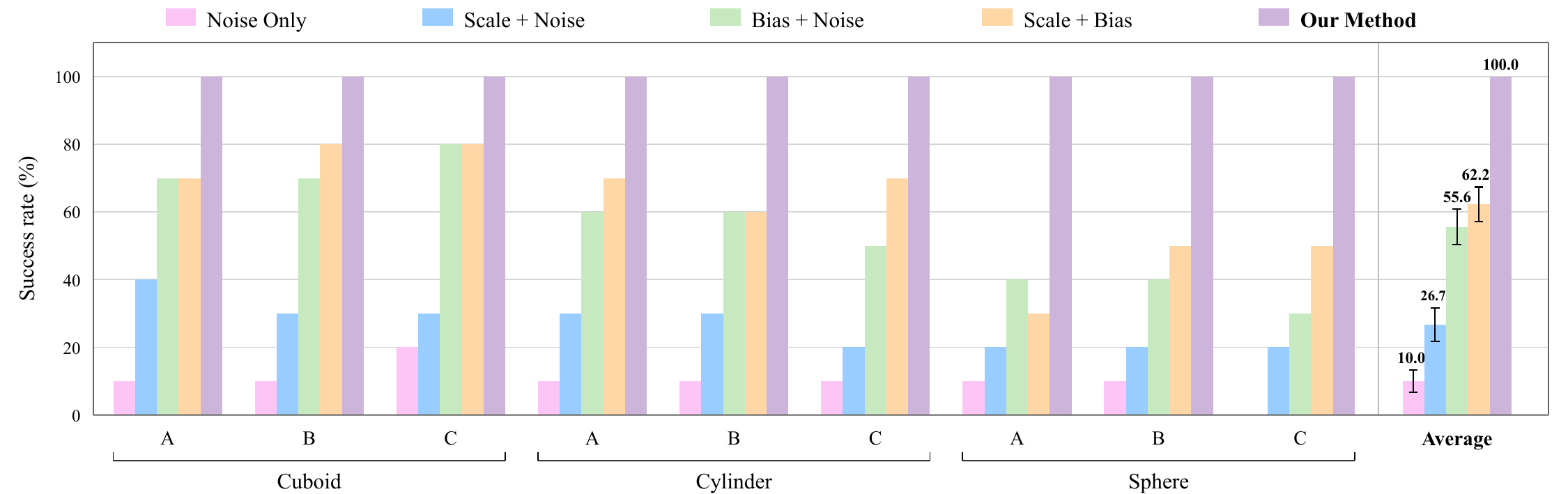}
\vspace{-7mm}
\caption{Real-world ablation of the three alignment operations on nine
in-distribution objects (10 trials per object and variant). Bars show
observed success rates. Error bars are shown only for Average and denote
$\pm 1$ standard error of the equally weighted mean across the nine
objects, computed by propagating the per-object sampling variances
under independent trials. Per-object error bars are omitted for clarity.}
\label{fig:ablation}
\vspace{-0.3cm}
\end{figure*}

This procedure exposes the policy to torque noise with a magnitude
determined from real measurements while applying the same temporal-differencing process in both domains.

\subsection{Learning Method}\label{sec:learn}
We consider a proprioceptive grasping task with a
direct-drive multifingered gripper. Due to the high mechanical
transparency of direct-drive actuation, contact-induced torque changes
can be observed through motor current without tactile or external force
sensors. The deployed policy therefore grasps, lifts, and holds an object using only proprioceptive observations (joint
position $q_t$ and joint-torque change $\Delta\tau_t$).
Training follows a teacher--student framework~\cite{chen2021reorientation,lee2026blind}, illustrated in Fig.~\ref{fig:learning_framework}.
The teacher policy uses a multilayer perceptron (MLP) with hidden dimensions [512, 256, 128].
We train the teacher using proximal policy optimization
(PPO)~\cite{ppo2017} in Isaac Lab~\cite{mittal2025isaaclab} with 9,000 parallel simulation
environments for 2,000 iterations on an RTX 5090 GPU. For each student-policy configuration, we collect 90,000
demonstration trajectories.
Teacher rollouts are retained when the task reward $r_t$
at the final step exceeds the corresponding selection threshold.
The student policy uses a Transformer trained by behavior cloning
for 2,000 epochs on the collected trajectories. The hyperparameters for teacher and
student training are summarized in
Table~\ref{tab:training_hyperparameters}. We use cuboids, cylinders, and spheres as training objects,
with three size variants (A--C) for each shape. Their dimensions
are listed in Table~\ref{tab:object_dimensions}.

The teacher receives joint-level proprioceptive
observations and simulator-only privileged information. Its observation
is defined as
\begin{equation*}
o^{\mathrm{T}}_t=
\big[
q_t,\Delta\tau_t,a_{t-1},\dot q_t,
\tau_{m,t},o^{\mathrm{priv}}_t
\big]\in\mathbb{R}^{82},
\end{equation*}
where $q_t$ and $\dot q_t$ denote the joint positions and velocities,
respectively; $\Delta\tau_t$ is the joint-torque difference defined
in \eqref{eq:delta}; $\tau_{m,t}$ is the joint-torque observation;
and $a_{t-1}$ is the previous action. The privileged observation
$o^{\mathrm{priv}}_t\in\mathbb{R}^{37}$ comprises the object position,
orientation, linear and angular velocities, fingertip positions and
orientations, and fingertip-to-object distances. In contrast, the deployable student receives only the joint positions
and joint-torque differences:
\begin{equation*}
o^{\mathrm{S}}_t=
\big[q_t,\Delta\tau_t\big]\in\mathbb{R}^{18}.
\end{equation*}
The action is a $9$-dimensional vector of relative joint-position
changes, $a_t=\Delta\theta_t\in\mathbb{R}^{9}$. The policy operates
at 20~Hz with a control period of $\Ts=50$~ms.

The reward function is designed to encourage smooth object lifting
while maintaining the object near the gripper center to promote
grasp stability. It combines a task reward with an action-magnitude
penalty and an action-rate penalty:
\vspace{-0.1cm}
\begin{equation}
r =
w_{1}r_t
+w_{2}r_a
+w_{3}r_{ar},
\label{eq:reward}
\end{equation}
\vspace{-0.5cm}

\noindent where $w_{1}=1.0$, $w_{2}=-0.004$, and
$w_{3}=-0.002$. The task reward is defined as
\vspace{-0.0cm}
\begin{equation}
r_t =
\begin{cases}
f_h(h_t)f_d(d_{xy}),
&(d_{xy}<d_{\max},\; h_t<h_{\mathrm{cut}})\\[3pt]
0 & (\text{otherwise})
\end{cases}
\label{eq:task_reward}
\end{equation}
\vspace{-0.1cm}

\noindent where $f_h$ measures the lifting progress and $f_d$ measures the
planar-centering progress:
\vspace{-0.1cm}
\begin{equation}
f_h(h_t)
=
1-
\left[
\operatorname{clip}
\left(
\frac{h^{\star}-h_t}{h^{\star}},0,1
\right)
\right]^{1/2},
\label{eq:height_reward}
\end{equation}
\vspace{-0.1cm}
\begin{equation}
f_d(d_{xy})
=
1-
\left[
\operatorname{clip}
\left(
\frac{d_{xy}-d_{\mathrm{tol}}}
{d_{\max}-d_{\mathrm{tol}}},0,1
\right)
\right]^{1/2}.
\label{eq:distance_reward}
\end{equation}
\vspace{-0.1cm}

\noindent Here, $h_t$ denotes the object height relative to its
environment-specific initial height, and $d_{xy}$ denotes the planar distance
between the object and the gripper center. We use
$h^{\star}=3\,\mathrm{cm}$,
$d_{\mathrm{tol}}=0.1\,\mathrm{cm}$,
$d_{\max}=10\,\mathrm{cm}$, and
$h_{\mathrm{cut}}=4\,\mathrm{cm}$. The initial height is recorded at the
first post-reset step of each simulation environment.

The action-magnitude penalty discourages large relative
joint-position commands and is defined as
\begin{equation}
r_a = \lVert a_t \rVert_2^2
= \sum_{j=1}^{9} a_{t,j}^{2}.
\label{eq:action_penalty}
\end{equation}
The action-rate penalty promotes smoother control during grasping
and lifting by penalizing abrupt changes between consecutive
commands:
\begin{equation}
r_{ar} = \lVert a_t-a_{t-1} \rVert_2^2
= \sum_{j=1}^{9}(a_{t,j}-a_{t-1,j})^{2}.
\label{eq:action_rate_penalty}
\end{equation}
An episode ends on timeout after $10$\,s, or when the object leaves the
workspace, defined as a planar displacement exceeding $10$\,cm from the
gripper center.

\begin{figure*}[t]
    \centering
    \includegraphics[width=\textwidth]{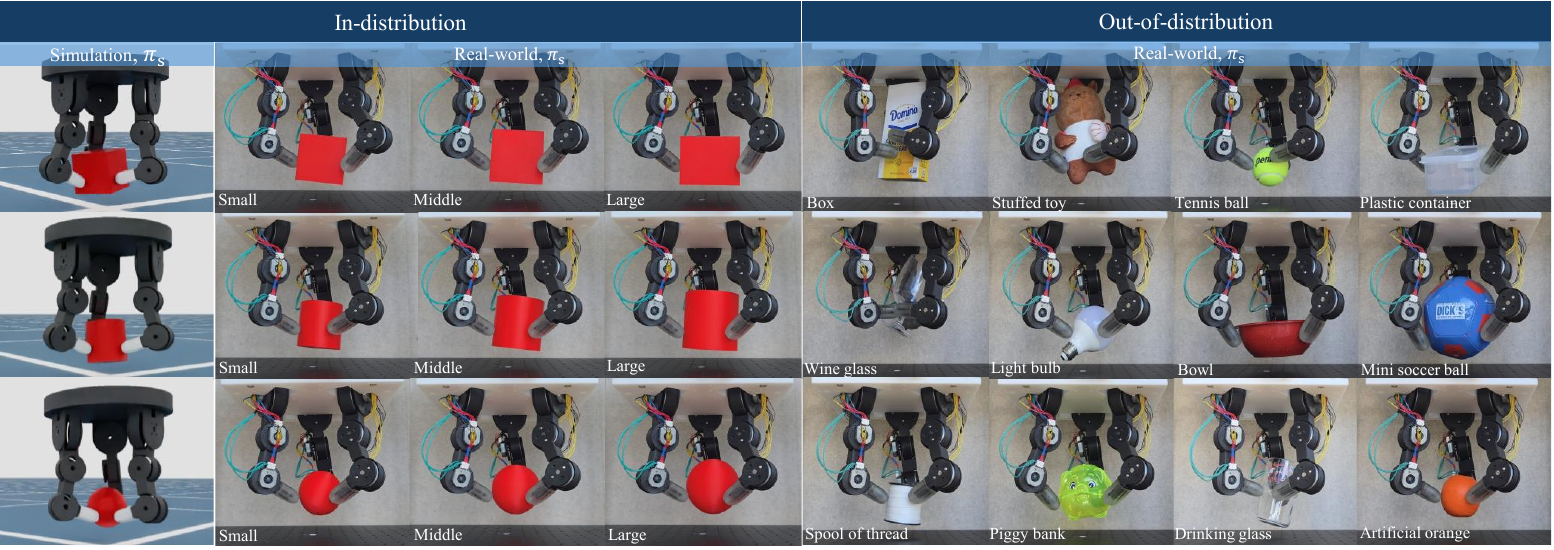}
    \vspace{-7mm}
    \caption{Objects used in the real-world evaluation. The evaluation
    includes 21 objects: nine in-distribution (ID) objects and twelve
    out-of-distribution (OOD) objects. The ID objects belong to the
    cuboid, cylinder, and sphere categories used during simulation
    training. The OOD objects are real-world items with previously unseen
    shapes, materials, and surface properties that fit within the gripper
    workspace.}
    \label{fig:real_objects}
    \vspace{-0.4cm}
\end{figure*}

We apply domain randomization to initial conditions, physical
parameters, and external disturbances (see Table~\ref{tab:events}).
Joint-angle offsets and object poses are resampled at each episode reset. Object mass, actuator stiffness and damping, and static and dynamic friction coefficients are randomized once at environment initialization and remain fixed across subsequent episodes. Downward velocity perturbations are applied to the object at
random time intervals to encourage stable grasping.

\section{Experiments}\label{sec:exp}
\begin{figure*}[t]
    \centering
    \includegraphics[width=\textwidth]{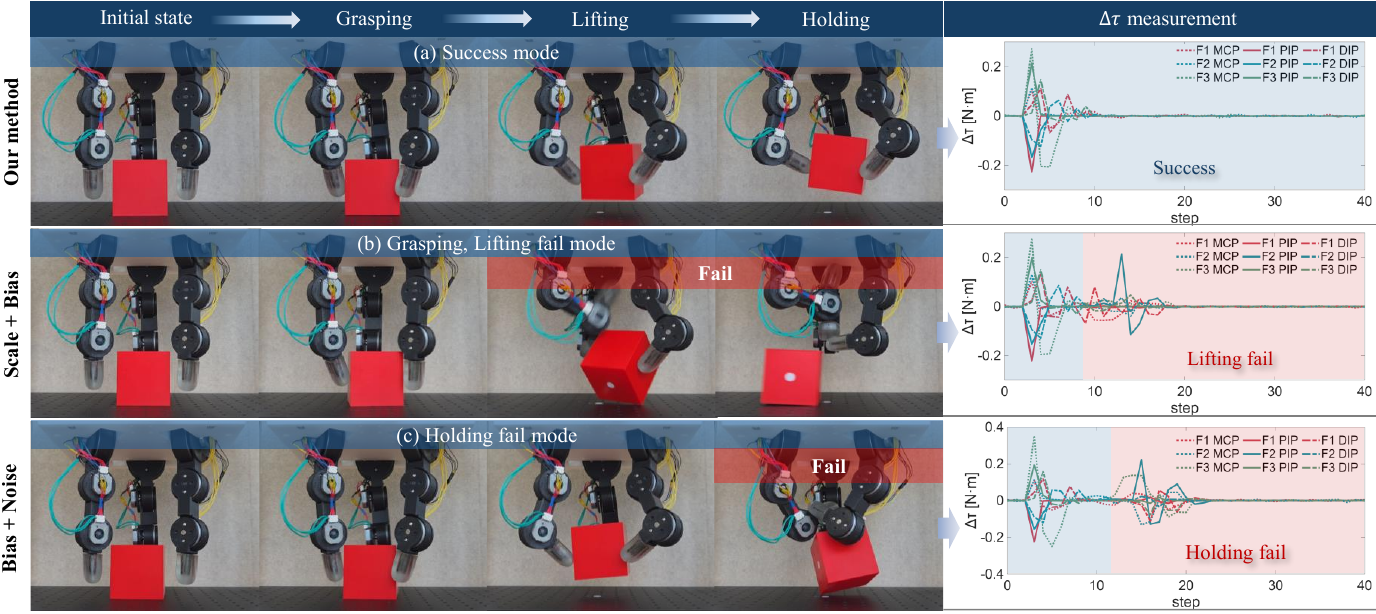}
    \vspace{-7mm}
    \caption{Representative real-world grasping sequences and motor-current-derived $\Delta\tau$ trajectories for the complete method and two $\Delta\tau$ ablations on the DD gripper. The sequences show grasp outcomes alongside the torque-change observations. \textcolor{black}{(a) \textbf{Our Method}, using calibrated
$K_\tau^\ast$ and noise injection, completes the grasp--lift--hold task.
(b) \textbf{Scale + Bias}, using calibrated $K_\tau^\ast$ without noise
injection, exhibits a grasping/lifting failure. (c) \textbf{Bias +
Noise}, using the datasheet $K_\tau$ with noise injection, exhibits a
holding failure. In the plot legends, F1--F3 denote the three fingers,
while MCP, PIP, and DIP denote the three joints of each finger.}}
    \label{fig:failure_modes}
    \vspace{-0.4cm}
\end{figure*}

This section evaluates the proposed observation alignment through
zero-shot grasping on a DD multifingered gripper using only actuator
proprioception. Each trial consists of four
stages: a prescribed non-contact initial state, grasping, lifting,
and holding. First, we compare five alignment variants on nine in-distribution (ID)
objects (Fig.~\ref{fig:ablation}). The
variants use different combinations of torque-constant calibration,
temporal differencing, and measurement-derived noise injection. This
experiment examines how each component affects real-world grasping
performance. Second, we evaluate our method on 21 objects. The evaluation includes
nine ID objects and twelve out-of-distribution (OOD) objects with unseen shapes, materials,
and surface properties. This experiment assesses grasping performance
and generalization to unseen objects (Fig.~\ref{fig:real_objects}). Third, we analyze representative grasping
sequences and
$\Delta\tau$ trajectories to characterize failure patterns during
grasp establishment, lifting, and holding (Fig.~\ref{fig:failure_modes}). All policies are trained entirely in simulation and transferred to the real-world without fine-tuning or adaptation.

\begin{table}[t]
\centering
\caption{Simulation and real-world success rates of student
policies $\pi_s$.}
\label{tab:student_sim_real}
\footnotesize
\setlength{\tabcolsep}{3pt}
\renewcommand{\arraystretch}{1.15}

\begin{tabular*}{\columnwidth}
{@{\extracolsep{\fill}}lcccc@{}}
\toprule
\multirow{2}{*}{Policy}
& \multirow{2}{*}{Observation}
& \multirow{2}{*}{Noise}
& \multicolumn{2}{c}{Success rate} \\
\cmidrule(lr){4-5}
& & & Simulation & Real-world \\
\midrule
\textcolor{black}{\textbf{Position Only}}
& $[q_t]$
& N/A
& 68.9\%
& 15.6\% \\

\textcolor{black}{\textbf{Scale + Noise}}
& $[q_t,\tau_t]$
& On
& 100.0\%
& 26.7\% \\

\textcolor{black}{\textbf{Scale + Bias}}
& $[q_t,\Delta\tau_t]$
& Off
& 100.0\%
& 62.2\% \\

\textbf{Our Method}
& $[q_t,\Delta\tau_t]$
& On
& 100.0\%
& \textbf{100.0\%} \\
\bottomrule
\end{tabular*}

\par\smallskip
{\scriptsize\raggedleft
All torque-based policies listed here use calibrated
$K_\tau^\ast$ at deployment.\par}
\vspace{-0.4cm}
\end{table}
\subsection{Ablation Study}\label{sec:ablation}


We evaluate five torque-observation alignment variants on nine ID objects. The objects match the
training shapes and dimensions listed in
Table~\ref{tab:object_dimensions}. We conduct ten trials per object for
each variant. Each trial starts from the same prescribed non-contact
initial condition. A trial is successful when the gripper grasps the
object, lifts it completely from the supporting surface, and holds it
until the end of the 10-s episode.

The five variants use different combinations of the three alignment
components. \textbf{Noise Only} uses the datasheet $K_\tau$ and absolute
torque $\tau$ with measurement-derived noise injection.
\textbf{Scale + Noise} replaces the datasheet value with the calibrated
$K_\tau^\ast$ while retaining absolute torque. \textbf{Bias + Noise} uses
the datasheet $K_\tau$ and temporal difference $\Delta\tau$ with noise
injection. \textbf{Scale + Bias} combines the calibrated $K_\tau^\ast$ and
$\Delta\tau$ without noise injection. \textbf{Our Method} combines the
calibrated $K_\tau^\ast$, $\Delta\tau$, and measurement-derived noise
injection.

We train three torque-based student-policy configurations: an absolute-torque policy
with noise, a $\Delta\tau$ policy with noise, and a $\Delta\tau$ policy
without noise. \textbf{Noise Only} and \textbf{Scale + Noise} use the
same absolute-torque policy, but one applies the datasheet $K_\tau$ and the
other applies the calibrated $K_\tau^\ast$ during deployment. Likewise, \textbf{Bias + Noise} and
\textbf{Our Method} use the same noise-injected $\Delta\tau$ policy.
\textbf{Scale + Bias} uses a $\Delta\tau$ policy trained without noise
injection. All other training and deployment settings were kept identical across the variants.
Table~\ref{tab:student_sim_real} summarizes the simulation and
real-world success rates of the student policies $\pi_s$. \textcolor{black}{Simulation evaluation also uses 10 trials per ID object,
for a total of 90 trials per policy.}
We include a \textcolor{black}{\textbf{Position Only}} policy to evaluate the contribution
of torque feedback beyond joint positions.
The \textcolor{black}{\textbf{Position Only}} student achieves 68.9\% success in simulation,
suggesting difficulty in reproducing the teacher's behavior
using joint positions alone, and 15.6\% in the real world.
All three torque-based students achieve 100.0\% success in simulation.
In the real world, the proposed method maintains 100.0\% success,
whereas the absolute-torque policy achieves 26.7\% and the
torque-difference policy without noise injection achieves 62.2\%.
These results support the value of torque feedback for
distillation and highlight the importance of observation
alignment for real-world deployment.

\begin{table}[t]
\centering
\caption{Real-world grasping success rates on ID and OOD objects.}
\label{tab:id_ood_results}
\scriptsize
\setlength{\tabcolsep}{4pt}
\renewcommand{\arraystretch}{0.92}

\begin{tabular*}{\columnwidth}{
@{\extracolsep{\fill}}
r l r
@{\hspace{1em}}
r l r
@{}
}
\toprule
\multicolumn{3}{c}{\textbf{ID Objects}} &
\multicolumn{3}{c}{\textbf{OOD Objects}} \\
\cmidrule(r){1-3}
\cmidrule(l){4-6}

\textbf{\#} &
\textbf{Object} &
\textbf{Rate} &
\textbf{\#} &
\textbf{Object} &
\textbf{Rate} \\
\midrule

1 & Cuboid A   & 100\% &
10 & Box               & 100\% \\

2 & Cuboid B   & 100\% &
11 & Stuffed toy       & 100\% \\

3 & Cuboid C   & 100\% &
12 & Tennis ball       & 90\% \\

4 & Cylinder A & 100\% &
13 & Plastic container & 100\% \\

5 & Cylinder B & 100\% &
14 & Wine glass        & 100\% \\

6 & Cylinder C & 100\% &
15 & Light bulb        & 90\% \\

7 & Sphere A   & 100\% &
16 & Bowl              & 100\% \\

8 & Sphere B   & 100\% &
17 & Mini soccer ball  & 100\% \\

9 & Sphere C   & 100\% &
18 & Spool of thread   & 100\% \\

& & &
19 & Piggy bank        & 100\% \\

& & &
20 & Drinking glass            & 100\% \\
& & &
21 & Artificial orange          & 100\% \\

\midrule
\multicolumn{2}{r}{\textbf{Average}} &
\textbf{100.0\%} &
\multicolumn{2}{r}{\textbf{Average}} &
\textbf{98.3\%} \\

\midrule
\multicolumn{5}{r}{\textbf{Overall Average}} &
\textbf{99.0\%} \\
\bottomrule
\end{tabular*}
\vspace{-0.6cm}
\end{table}

\textbf{Our Method} achieves 100.0\% success on all nine ID objects, as
shown in Fig.~\ref{fig:ablation}. \textbf{Noise Only} achieves 10.0\%,
and \textbf{Scale + Noise} achieves 26.7\%. These results indicate that the remaining bias mismatch significantly
affects reliable sim-to-real transfer of absolute-torque observations. \textbf{Bias + Noise} increases the success rate to
55.6\%, showing the benefit of temporal differencing. Adding scale
calibration increases this result to 100.0\% with \textbf{Our Method}. This 44.4-percentage-point improvement highlights the
importance of torque-constant calibration for reliable
zero-shot policy transfer. \textbf{Scale + Bias} achieves 62.2\%, whereas adding noise injection
increases the result to 100.0\%. This 37.8-percentage-point improvement supports the use
of noise injection during training to improve real-world
grasping performance. Together, these
comparisons demonstrate the complementary roles of all three alignment
operations in zero-shot sim-to-real transfer.

\subsection{Real-World Grasping Experiments}\label{sec:real_world}
We evaluate our method on 21 objects under the same protocol as the
ablation study. The twelve OOD objects are previously unseen
real-world objects with combinations of geometry, material, and surface
properties not represented during training. Each object is evaluated
over 10 trials from the same prescribed initial condition.

As summarized in Table~\ref{tab:id_ood_results}, our method achieves
average success rates of 100.0\% on the ID objects and 98.3\% on the OOD
objects, resulting in an overall success rate of 99.0\%. The two unsuccessful trials involved the light bulb and the
tennis ball. The light bulb tilted to one side before falling
during holding, whereas the tennis ball slipped between the fingers during lifting. These results show that our method enables zero-shot proprioceptive grasping of previously unseen objects using DD actuator proprioception. The high mechanical transparency of DD actuation allows motor-current-derived torque changes to provide cues about contact onset and changing interaction loads. The deployed policy only uses these cues together with joint positions, without external sensors.

\subsection{Failure-Mode Analysis}\label{sec:failure_analysis}
We examine representative unsuccessful trials to illustrate
failure patterns during grasp establishment, lifting, and holding.
Because the absolute-torque variants do not achieve reliable transfer,
as shown in Fig.~\ref{fig:ablation}, the analysis focuses on the three
$\Delta\tau$ variants in Fig.~\ref{fig:failure_modes}. These variants
are the \textcolor{black}{\textbf{Our Method}}, the
calibrated variant without training noise \textcolor{black}{\textbf{Scale + Bias}}, and
the datasheet-$K_\tau$ variant \textcolor{black}{\textbf{Bias + Noise}}.

All three variants exhibit an initial torque transient during closure.
After contact, the complete method produces a coordinated torque
transition followed by rapid stabilization. In contrast, both ablated
variants exhibit continued torque redistribution after the initial
closure transient.
The datasheet values overestimate the GL60 and GL40 torque channels by
38.0\% and 7.8\%, respectively, relative to their calibrated values.
These unequal scale errors alter the relative torque pattern across
joints driven by the two motor types. The corresponding grasp sequence
shows continued closure after contact, middle-finger overtravel, and
asymmetric support by the remaining fingers. The variant trained without
measurement-derived noise instead exhibits repeated action corrections
and unstable torque redistribution before lifting.

Both observation mismatches are associated with a loss of balanced
three-finger support. Grasp-establishment failure occurs when a stable
three-finger contact configuration is not formed. Lift failure occurs
when an asymmetric grasp cannot support the object during lifting, while
hold failure occurs when the object is lifted but subsequently loses
support. In contrast, \textbf{Our Method} stabilizes the grasp after contact and
maintains coordinated support throughout grasping, lifting, and holding.
Together, the ablation results and measured $\Delta\tau$ trajectories
show that the proposed torque-observation alignment improves the reliability of zero-shot sim-to-real transfer.

\section{Discussion and Limitations}\label{sec:disc}
Despite the successful sim-to-real transfer, the proposed method
has several limitations. First, the
deployed policy uses torque differences rather than absolute torque
and does not directly observe sustained torque levels. \textcolor{black}{In our task, lifting involves dynamic motion, whereas the
settled holding phase is quasi-static. Baseline suppression
assumes slowly varying offsets, so more dynamic tasks may
require additional compensation.}
Second, the torque constants and noise parameters are identified
under specific dynamometer conditions. Their validity across motor
units and operating temperatures requires further evaluation.
Third, the experiments focus on single-object grasping with a fixed-base gripper under prescribed initial conditions, using objects within its workspace. Further evaluation with an arm-mounted gripper is needed to assess grasping across a wider variety of objects.

The results highlight the value of motor-current feedback for
sim-to-real grasping when scale, baseline, and noise differences
are addressed. The proposed method aligns policy observations
without learning an actuator model or requiring real-world policy
training. The resulting policy grasps unseen objects using joint
positions and torque changes, without vision or dedicated tactile
and force/torque sensors. These findings support proprioceptive
feedback as a useful source of contact information for DD grippers.

Future work will integrate the DD gripper with a robotic arm
to evaluate grasping across a broader range of objects with
different sizes and stiffnesses, and investigate damage-free
grasping of deformable objects using proprioceptive torque
feedback without external sensors.
\section{Conclusion}

We presented a simple torque-observation alignment method for
robots with DD actuators and validated it through zero-shot
grasping on a DD multifingered gripper.
Dynamometer calibration reduces torque-scale mismatch,
temporal differencing cancels constant offsets and attenuates
slowly varying baselines, and Gaussian noise based on
dynamometer residuals is injected during training to improve
robustness to residual observation noise. The method requires
neither a learned actuator model nor real-world policy training.

Real-world ablations show the highest success when all three
alignment components are combined. The policy achieved 100.0\% success on nine ID objects and
98.3\% on twelve OOD objects, with an overall success rate
of 99.0\%. Building on the high mechanical transparency of DD actuation, the proposed approach uses simple observation processing to make motor-current-derived torque changes usable as feedback about contact onset and varying interaction loads. Under the evaluated grasp--lift--hold conditions, joint
positions and torque differences support zero-shot
proprioceptive grasping without vision or dedicated tactile
or force/torque sensors.

\bibliographystyle{IEEEtran}
\bibliography{References}
\end{document}